\documentclass[twocolumn,10pt]{article}
\usepackage[top=2.5truecm,bottom=2.54truecm,left=2.0truecm,right=2.0truecm]{geometry}
\usepackage{newtxtext,newtxmath}
\usepackage{latexsym}
\usepackage{graphicx}

\newcommand{\argmin}{\mathop{\rm arg~min}\limits}
\usepackage{amsmath, amssymb}
\usepackage{type1cm}
\usepackage{comment}
\usepackage{authblk}
\usepackage{cite}
\makeatletter
\renewcommand{\section}{%
\@startsection{section}{1}{\z@}%
{-1.5ex \@plus -1ex \@minus -.2ex}%
{1.7ex \@plus.2ex}%
{\large \bfseries}}

\makeatother
\makeatletter
\renewcommand{\subsection}{%
\@startsection{subsection}{1}{\z@}%
{-1ex \@plus -1ex \@minus -.2ex}%
{1ex \@plus.2ex}%
{\large \bfseries}}

\makeatother
\usepackage{caption}
\usepackage{indentfirst}
\newcommand\figref[1]{\textbf{Fig.~\ref{#1}}}
\newcommand\tabref[1]{\textbf{Tab.~\ref{#1}}}

\begin{document}
\twocolumn[
\begin{center}
\fontsize{14pt}{14pt}\textbf{Convolutional Nonlinear Dictionary with Cascaded Structure Filter Banks}\\ \vspace{24pt}
\fontsize{12pt}{12pt}Ruiki KOBAYASHI\hspace{1cm}Shogo MURAMATSU\\
\textit{Dept. of Electrical and Electronic Eng., Niigata Univ.}\\
\textit{shogo@eng.niigata-u.ac.jp}\\
\vspace{12pt}
\end{center}
]

\begin{center}
    \textbf{\large Abstract}
\end{center}

\textit{This study proposes a convolutional nonlinear dictionary (CNLD) for image restoration using cascaded filter banks. Generally, convolutional neural networks (CNN) demonstrate their practicality in image restoration applications; however, existing CNNs are constructed without considering the relationship among atomic images (convolution kernels). As a result, there remains room for discussing the role of design spaces. To provide a framework for constructing an effective and structured convolutional network, this study proposes the CNLD. The backpropagation learning procedure is derived from certain image restoration experiments, and thereby the significance of CNLD is verified. It is demonstrated that the number of parameters is reduced while preserving the restoration performance.}

\section{Introduction}
With the development of measurement technologies, it is now possible to acquire a large amount of different physical data, such as tomographic imaging data and multi-dimensional time-series data. Simultaneously, the demand for high performance signal restoration is increasing. For high-quality signal restoration, it is advisable to use a generative model that can effectively represent the original signal. The generative model is a mathematical expression of the prior knowledge about the target signal. Sparse modeling is employed in few image restoration techniques based on a generative process, wherein the generative model assumes that the essential information of the target signal is sparse. The framework of the generative model is provided by a dictionary comprising atomic waveforms, whose role is to synthesize signals from low-dimensional features. However, linear dictionaries are incompatible with nonlinear signal processing structures.


In recent years, artificial intelligence (AI) has advanced remarkably developed. In particular, deep learning, such as convolutional neural networks (CNNs), has been applied in various fields, including signal restoration \cite{Alex,Zhang}. CNN is a feed-forward-type neural network involving multiple layers of convolution and nonlinear functions that extract the local features of signals. A conventional CNN can take the role of a generative model, but it requires a large amount of data for training. It learns not only the generative process but also the observation process. A large memory capacity is required for the large number of design parameters of a CNN, and overtraining is likely to occur if there is insufficient training data. Deep Image Prior adapts a generative model explicitly separated from the observation model based on sparse modeling \cite{Ulyanov}. Therefore, prior knowledge about observation and generation can be reflected in the network architecture. However, it requires a high degree of freedom in the network architecture to learn the signal generation process from random number inputs.

This study proposes a nonlinear dictionary to solve the design problems in CNNs that rely on a large amount of training data and image restoration problems in sparse modeling. To construct a nonlinear dictionary, we propose to utilize filter banks whose adjoint operators are clear. For filter banks, we have adequate knowledge on the structure and can reflect prior knowledge to the network architecture. In this study, to provide a framework for the construction of an effective nonlinear dictionary, we propose a convolutional nonlinear dictionary (CNLD), where an activation function provides the building blocks of cascaded filter banks and derives its learning methods.

\section{Overview of Image Restoration}
This section provides an overview of the image restoration by sparse modeling and by CNNs.

Sparse modeling assumes that the essential information of a signal is sparse. By using an appropriate dictionary $\mathbf{D}$, a signal $\mathbf{x}$ can be sparsely approximated or represented. The synthesis of $\mathbf{x}$ is expressed as $\mathbf{x}=\mathbf{Dy}$, where $\mathbf{y}$ is a coefficient vector. The coefficient vector $\mathbf{y}$ is expressed as
\begin{equation}
 \hat{\mathbf{y}} = \argmin_{\mathbf{y}} \|\mathbf{x} - \mathbf{D} \mathbf{y}\|_2^2 + \lambda \|\mathbf{y}\|_1,
\label{eq:sparse} 
\end{equation}
where $\mathbf{D}\in{\mathbb{R}}^{N\times M}$,  $\mathbf{x}\in{\mathbb{R}}^M$,  $\|\cdot\|_2$ is the standard norm, $\|\cdot\|_1$ is $\ell_1$norm, $M$ is the number of elements in $\mathbf{y}$, $N$ is the number of elements in $\mathbf{x}$, and $\lambda$ is the regularization parameter.
\eqref{eq:sparse} denotes the least absolute shrinkage and selection operator (LASSO) in \cite{Robert}. The optimization of the problem in \eqref{eq:sparse} by the proximity gradient method is known as the iterative shrinkage thresholding algorithm (ISTA).
The signal $\mathbf{x}$ can be sparsely expressed; i.e., any noise that do not follow the feature can be removed.
However, existing linear dictionaries cannot represent nonlinear signal generation processes efficiently.

\begin{figure}[tb]
\centering
        \includegraphics[width = 0.95\linewidth]{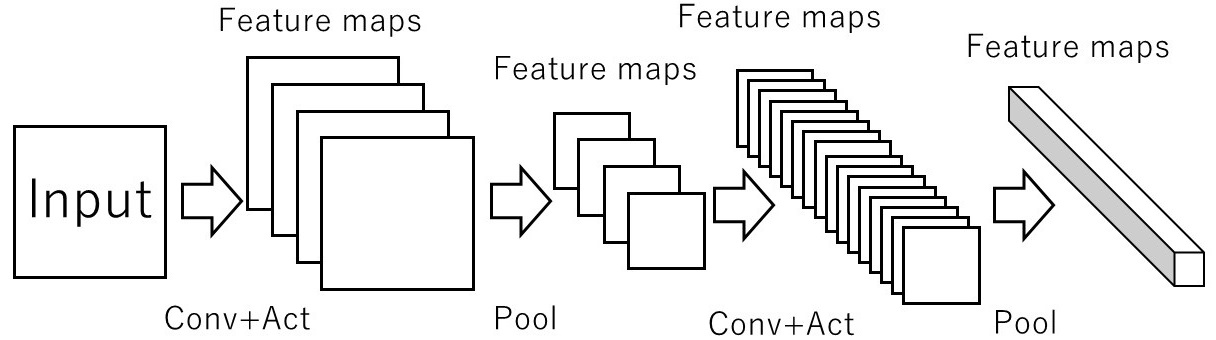}
\caption{Example architecture of CNN for feature extraction, where ``conv'' is convolution by filter, ``Act'' is nonlinear activation, and ``Pool'' is pooling.}\label{fig:CNN}
\end{figure}
\figref{fig:CNN} depicts an example of the architecture of CNN \cite{Lecun}. The CNN extracts local features by combining dimensionality reduction by operations, such as convolution, order statistics, downsampling, and nonlinear activation. It is also possible to recognize and restore signals from the extracted features.

For CNN, various modifications are developed, and some are applied to image denoising techniques, such as in DnCNN \cite{Zhang}; however, DnCNN has few unresolved issues. Due to the large number of design parameters, overfitting is prone to occur in the absence of a large amount of training data. Further, unnecessary design parameters may exist if proper design spaces, such as symmetry and orthonormality of atomic waveforms, are not considered.


\section{Overview of Filter Banks}
In this section, we review the filter banks and their role in image restoration. Filter banks comprises of analysis and synthesis systems of signals with various filters, including downsamplers and upsamplers. They have a clear relation such that the analysis and synthesis systems are adjoint to each other when the filters are in the flipped relationship. A substantial amount of information regarding the design of filter banks has been accumulated thus far. Therefore, it may be useful to structurally reflect our knowledge to generate the target signals.
\begin{figure}[tb]
\centering
        \includegraphics[width=0.99\linewidth]{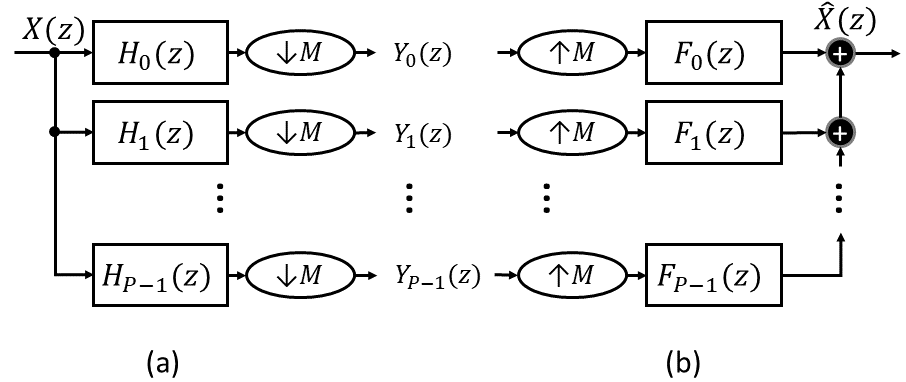}
\caption{Filter bank structure. (a) analysis filter bank and (b) synthesis filter bank, where $X(z)$ is the input signal, $Y_p(z)$ is $p$-th coefficient signal, $\hat{X}(z)$ is the output signal, $H_p(z)$ is $p$-th analysis filter, $F_p$ is $p$-th synthesis filter, and $\downarrow M$ and $\uparrow M$ are downsampling and upsampling with factor $M$, respectively.}
\label{fig:Fil}
\end{figure}
\figref{fig:Fil} depicts an example structure of an analysis--synthesis system with $P$-channel uniformly 
decimated filter banks, where $M$ is the downsampling factor. In the figure, $X(z)$ is the input signal, $ \hat{X}(z)$ is the output signal, and $\downarrow M$ and $\uparrow M$ are the downsampler and upsampler with factor $M$, respectively.

Filter banks as depicted in \figref{fig:Fil} can be designed and implemented efficiently by using polyphase representations. \figref{fig:Fil} can be represented by polyphase matrices $\mathbf{E}(z)\in\mathbb{R}^{P\times M}$ and $\mathbf{R}(z)\in\mathbb{R}^{M\times P}$. The analysis filter bank $\{H_p(z)\}^{P-1}_{p=0}$ with factor $M$ is expressed as
\begin{equation}
    \mathbf{h}(z) = (H_0(z),H_1(z),\cdots ,H_{P-1}(z))^{\intercal} = \mathbf{E}(z^M)\mathbf{d}(z),
\end{equation}
and the synthesis filter bank $\{F_p(z)\}^{P-1}_{p=0}$ with factor $M$ is expressed as
\begin{equation}
    \mathbf{f}^\intercal(z) = (F_0(z),F_1(z),\cdots ,F_{P-1}(z)) = \mathbf{d}^\intercal(z^{-1})\mathbf{R}(z^M),
\end{equation}
where $[\mathbf{d}(z)]_\ell = z^{-\ell}$, $\ell \in \{0,1, and \cdots M-1\}$ is the delay chain vector. In the case of cascaded filter banks, the polyphase matrix can be factored into a product form, such as
 \begin{equation}
    \mathbf{R}(z) =\mathbf{R}_I(z)\mathbf{R}_{I-1}(z)\cdots\mathbf{R}_1(z)  =\prod_{i = 1}^{I}\mathbf{R}_i(z).
    \label{eq:cascade}
\end{equation}
Parameterization of the design by some additional structural constraints to the $\mathbf{R}_i(z)$ provides an effective reduction of the design space.

The impulse response of each filter gives an atomic element (atom), and a set of atoms constitutes a dictionary.
To optimize the signal representation of the dictionary $\mathbf{D}$, it is effective to train the atoms that match a set of target signals. The problem setting for dictionary learning is represented by
\begin{align}
    \{\hat{\mathbf{D}},\{\hat{\mathbf{y}}_j\}\} = \argmin_{\mathbf{D},\{\mathbf{y}_j\}}\frac{1}{2S}\sum_{j=1}^{S} \|\mathbf{x}_j - \mathbf{D} \mathbf{y}_j\|_2^2, \notag \\ \text{s.t. }\ \|\mathbf{y}_j\|_0 \leq K, j \in \{ 1, 2, \cdots S \},
    \label{eq:problem}
\end{align}
where $\mathbf{y}_j\in\mathbb{\mathbb{R}}^N$ is a coefficient vector for the $j$-th sample $\mathbf{x}_j\in\mathbb{\mathbb{R}}^M$ and $S\in\mathbb{N}$ is the number of samples.
If the dictionary $\mathbf{D}$ is represented as a cascaded filter bank as in \eqref{eq:cascade}, it can be factored as 
\begin{equation}
    \mathbf{D}_{\boldsymbol{\theta}} = \mathbf{F}_I\mathbf{F}_{I-1} \cdots \mathbf{F}_1,
\end{equation}
where $\mathbf{F}_i$ is the global matrix representation corresponding to the polyphase factor $\mathbf{R}_i(z)$ and $\boldsymbol{\theta}$ is the set of design parameters. For optimizing the design parameters in $\boldsymbol{\theta}$, a loss function can be defined as the sum of squared errors of the generated and training signals. The loss function  of the $j$-th sample $J_j(\boldsymbol{\theta})$ is represented as
\begin{equation}
        J_j(\boldsymbol{\theta}) = \frac{1}{2}\|\mathbf{x}_j - \mathbf{D}_{\boldsymbol{\theta}} \mathbf{y}_j\|_2^2 = \frac{1}{2}\|\mathbf{r}_{\mathbf{x}_j}(\boldsymbol{\theta})\|_2^2,
    \label{eq:error}
\end{equation}
where $\mathbf{r}_{\mathbf{x}_j}(\boldsymbol{\theta})$ is the error function of the $j$-th sample. Let us consider minimizing $J_j(\boldsymbol{\theta})$.
The partial derivative $\frac{\partial J_j}{\partial \theta_i}$ of the loss function $J_j(\boldsymbol{\theta})$ with the $i$-th parameter is written as 
\begin{align}
    \frac{\partial J_j}{\partial \theta_i} 
    &= -\Big\langle \mathbf{F}_{i+1}^{\intercal}\mathbf{F}_{i+2}^{\intercal}\cdots \mathbf{F}_{I}^{\intercal}\mathbf{r}_{\mathbf{x}_j}(\boldsymbol{\theta}), 
    \left( \frac{\partial \mathbf{F}_i}{\partial \theta_i} \right) \mathbf{F}_{i-1} \cdots \mathbf{F}_1\mathbf{y}_j
    \Big\rangle, \label{eq:der}
\end{align}
where $\langle \cdot, \cdot \rangle$ represents the inner product. 
Without loss of generality, we simply assume that each building block $\mathbf{R}_i({z})$ has one independent parameter $\theta_i$.
The first argument of the inner product in \eqref{eq:der} can be calculated by the backpropagation manner until the parameters are independent on other factors. By using the result of this partial differentiation, the gradient $\nabla_{\boldsymbol{\theta}}J_j(\boldsymbol{\theta})$ is derived by
\begin{equation}
    \nabla_{\boldsymbol{\theta}}  J_j(\boldsymbol{\theta}) =
    \left( 
    \frac{\partial J_j}{\partial \theta_1},\frac{\partial J_j}{\partial \theta_2},\dots,\frac{\partial J_j}{\partial \theta_I}
    \right)^{\intercal}.
    \label{eq:grad}
    \end{equation}
Typical dictionary learning methods alternatively apply a sparse-approximation and dictionary update step.
The design parameters in $\boldsymbol{\theta}$ are updated in the dictionary update step by using this gradient.

\section{Convolutional Nonlinear Dictionary}
\begin{figure}[tb]
\centering
        \includegraphics[width=0.99\linewidth]{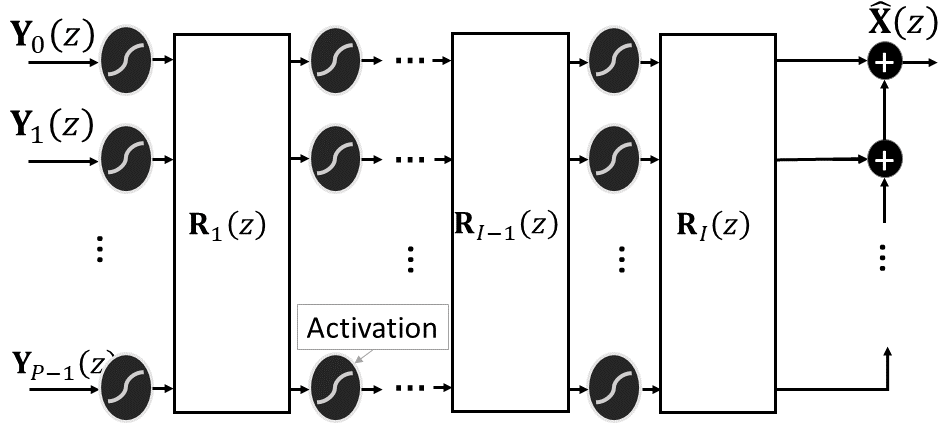}
\caption{Proposed structure of CNLD $\mathbf{\phi}_{\boldsymbol{\theta}}(\cdot)$ (Synthesizer)}
\label{fig:NlFil}
\end{figure}

The CNN and cascaded filter banks have some similarities. By constructing a CNN-like network using cascaded filter banks, we exploit the advantages of both systems. In this section, a framework for a structured CNLD is demonstrated. \figref{fig:NlFil} depicts an example configuration of our proposed CNLD.
As depicted in \figref{fig:NlFil}, CNLD $\mathbf{\phi}_{\boldsymbol{\theta}}:\mathbb{R}^N \xrightarrow{} \mathbb{R}^M$ is constructed by putting an activation scalar function $\sigma_i(\cdot)$ between the successive module of a cascaded filter bank as in \eqref{eq:cascade}. CNLD $\mathbf{\phi}_{\boldsymbol{\theta}}(\cdot)$ is expressed as
\begin{equation}
   \mathbf{\phi}_{\boldsymbol{\theta}}(\cdot) = f_I \circ f_{I-1} \circ \cdots \circ f_1(\cdot),
\end{equation}
where $f_i(\mathbf{x}) = \mathbf{F}_i\sigma_i(\mathbf{x})$. As in the existing CNNs, $\mathbf{\phi}_{\boldsymbol{\theta}}(\cdot)$ realizes a mapping that can be decomposed into a set of functions $\{f_i(\cdot)\}_{i=1}^I$ with independent design parameters. If the adopted filter bank satisfies the perfect reconstruction and the activation functions are invertible, $\mathbf{\phi}_{\boldsymbol{\theta}}$ is guaranteed to be invertible. The design parameters in $\boldsymbol{\theta}$ can be reduced appropriately. If the original filter bank is redundant, i.e., $M<N, \mathbf{\phi}_{\boldsymbol{\theta}}$ can be mapped from a manifold in $\mathbb{R}^N$. 

Let us consider training a CNLD $\mathbf{\phi}_{\boldsymbol{\theta}}$. For a given training set $\{\mathbf{x}_i\}_{i=1}^I$, we can formulate the design problem in the sparsity-aware dictionary learning manner as
\begin{align}
\{\hat{\boldsymbol{\theta}},\{\hat{\mathbf{y}}_j\}\} = \argmin_{\boldsymbol{\theta},\{\mathbf{y}_j\}}\frac{1}{2S}
\sum_{j=1}^{S}\|\mathbf{x}_j - \mathbf{\phi}_{\boldsymbol{\theta}}(\mathbf{y}_j) \|_2^2 ,
\notag \\
\text{s.t.}\ \|\mathbf{y}_j\|_0 \leq K, j\in\{1,2,\cdots,S\}.
\label{eq:nlproblema}
\end{align}
The loss function $J_j(\boldsymbol{\theta})$ of the $j$-th sample $\mathbf{y}_j$ is expressed by
\begin{align}
\mathbf{J}_j(\boldsymbol{\theta}) = \frac{1}{2}\|\mathbf{x}_j - \mathbf{\phi}_{\boldsymbol{\theta}}(\mathbf{y}_j) \|_2^2\ = \frac{1}{2}\|\mathbf{r}_{\mathbf{x}_j}(\boldsymbol{\theta}) \|_2^2 
\label{eq:nlproblemb}
\end{align}
as in \eqref{eq:error}.

The partial derivative $\frac{\partial J_j}{\partial \theta_i}$ of this loss function $J_j(\boldsymbol{\theta})$ is expressed as
\begin{align}
\frac{\partial J_j}{\partial \theta_i} 
&= -\Big\langle \mathbf{r}_{\mathbf{x}_j}(\boldsymbol{\theta}),\frac{\partial\mathbf{\phi}_{\boldsymbol{\theta}}(\mathbf{y}_j)}{\partial \theta_i} \Big\rangle \notag \\ 
&= -\Big\langle \mathbf{r}_{\mathbf{x}_j}(\boldsymbol{\theta}),\frac{\partial}{\partial \theta_i}\left( f_I \circ f_{I-1} \circ \cdots \circ f_1(\mathbf{y}_j) \right) \Big\rangle \notag \\ 
&= -\Big\langle\left(\frac{\partial f_{i+1}}{\partial f_{i}}\right)^{\intercal}\!\left(\frac{\partial f_{i+2}}{\partial f_{i+1}}\right)^{\intercal}\!\cdots\left(\frac{\partial f_I}{\partial f_{I-1}}\right)^{\intercal}\! \mathbf{r}_{\mathbf{x}_j}(\boldsymbol{\theta}), \notag \\
& \frac{\partial f_i}{\partial \theta_i} \circ f_{i-1} \circ \cdots f_1(\mathbf{y}_j)\Big\rangle
\end{align}
where $\frac{\partial J_j}{\partial \theta_i}$ is analyzed by the chain rule and can be calculated by the backpropagation method as long as the parameters are independent of each other and $\frac{\partial f_i}{\partial f_{i-1}}$ is a Jacobian matrix. The gradient is given as in \eqref{eq:grad}.
Thus, the design parameters in $\boldsymbol{\theta}$ for CNLD $\mathbf{\phi}_{\boldsymbol{\theta}}$ can be optimized through a gradient method.

\section{Performance Evaluation}
In this section, image denoising is performed to confirm the effectiveness of the proposed CNLD. We compare the performance of CNLD with that of DnCNN.

\begin{figure}[tb]
\centering
        \includegraphics[width=0.99\linewidth,height=55mm]{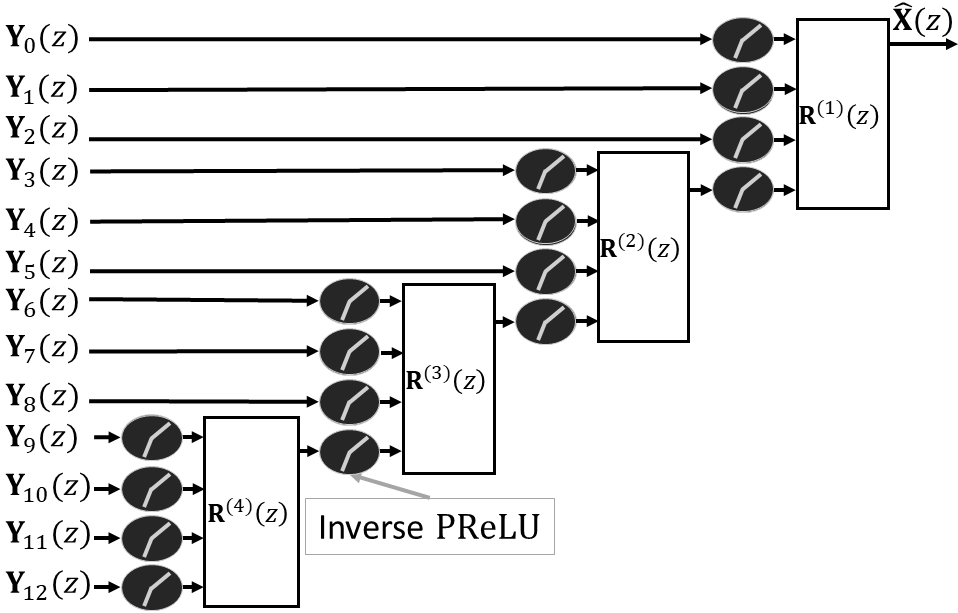}
\caption{CNLD structure used in the experiments (Synthesizer), where  a 2-dimensional separable CQF bank of polyphase order 1 is adopted for $\mathbf{R}^{(\ell)}(z)$. }\label{fig:nlcqf}
\end{figure}

\figref{fig:nlcqf} depicts the structure of CNLD used in this experiment, and \tabref{tab:cqf} illustrates the specifications of the adopted methods used for comparison.
\begin{table}[tb]
\caption{Experimental specifications}\label{tab:cqf}
  \centering
  \begin{tabular}{|c||c|c|c|} \hline
    Network     &DnCNN  &UDHT   &CNLD \\ \hline\hline
    Channels    &64     &13     &13 \\ \hline
    Layers      &20     &4      &4 \\ \hline
    Parameters  &668225 &0      &20 \\ \hline
  \end{tabular}
\end{table}
A conjugate quadrature filter (CQF) bank is used as the base filter bank for CNLD \cite{Vaid}. The undecimated construction is adopted to make it redundant. The CQF bank has a cascaded structure whose building block is defined as 
\begin{equation}\label{eq:cqf}
    \mathbf{R}_i(z) = \left(\begin{matrix}
      \cos{\theta_i} & -\sin{\theta_i}\\
      \sin{\theta_i} & \cos{\theta_i}\\
    \end{matrix} 
\right)
\left( \begin{matrix}
      z^{-1} & 0\\
      0 & 1\\
    \end{matrix} 
\right),
\end{equation}
where $\theta_i$ is the $i$-th design parameter. The CQF bank used in this experiment includes the undecimated Haar transform (UDHT) as a special case. The parametric rectified 
linear unit (PReLU) is used as the activation function \cite{he}, where the subdifferential is utilized instead of the differential to get a sub-gradient of the loss function. \tabref{tab:exam} summaries the learning specifications in this experiment.
\begin{table}[tb]
\caption{Learning specifications}\label{tab:exam}
  \centering
  \begin{tabular}{|c||c|} \hline
    Epoch & 10 \\ \hline
    Number of cases & 5, 10, 25 \\ \hline 
    Number of patches & 1280, 2560, 6400 \\\hline
    Minibatch size & 128 \\\hline
    AWGN standard deviation& $30/255$ \\\hline
  \end{tabular}
\end{table}
The performance of DnCNN with the specifications in the \tabref{tab:cqf} is also evaluated as a reference. The UDHT and CNLD are applied to the image denoising involving ISTA.

\begin{figure}[tb]
\centering
    \begin{minipage}[b]{0.24\linewidth}
        \centering
        \includegraphics[width = 0.99\linewidth]{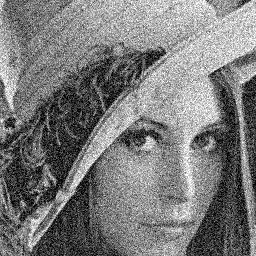}
        \centerline{\small (a) 18.62dB}
    \end{minipage}
    \begin{minipage}[b]{0.24\linewidth}
    \centering
        \includegraphics[width = 0.99\linewidth]{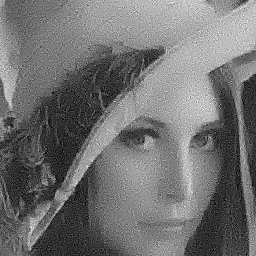}
        \centerline{\small (b) 22.14dB}
    \end{minipage}
    \begin{minipage}[b]{0.24\linewidth}
        \centering
        \includegraphics[width = 0.99\linewidth]{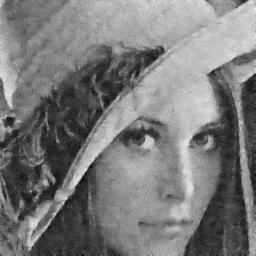}
        \centerline{\small (c) 26.14dB}
    \end{minipage}
        \begin{minipage}[b]{0.24\linewidth}
    \centering
        \includegraphics[width = 0.99\linewidth]{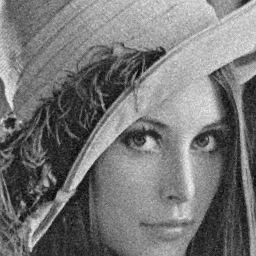}
        \centerline{\small (d) 27.71dB}
    \end{minipage}
\caption{Denoising results. (a) noisy image PSNR: 18.62dB, (b) UDHT PSNR: 22.14dB, (c) DnCNN PSNR: 26.14dB, and (d) CNLD PSNR: 27.71dB.}\label{fig:dn}
\end{figure}

\figref{fig:dn} and \tabref{tab:dn} illustrates the denoised results. Use peak signal-to-noise ratio (PSNR) as an evaluation index.
\begin{table}[tb]
\caption{Denoising results in PSNR, where the number in braces denoted is the number of training images.}    \label{tab:dn}
  \centering
  \begin{tabular}{|c||c|c|c|c|} \hline
&{\it Lena}&{\it Monar.}&{\it Barbara}&{\it Airplane}\\ \hline\hline
UDHT    &22.14&20.77&21.63&21.38\\ \hline
DnCNN   (5)&25.86&25.60&24.40&24.99\\ \hline
DnCNN   (10)&26.14&26.72&24.90&25.86\\ \hline
DnCNN   (25)&27.79&\textbf{27.33}&25.99&26.76\\ \hline
CNLD    (5)&26.90&26.29&26.85&26.54\\ \hline
CNLD    (10)&27.71&27.04&27.65&27.32\\ \hline
CNLD    (25)&\textbf{27.96}&27.30&\textbf{27.90}&\textbf{27.55}\\ \hline
  \end{tabular}
\end{table}
When compared with UDHT, the CNLD has approximately 5dB higher PSNR compared to that of UDHT. The effect of the learning design and nonlinear expansion is confirmed. When compared with DnCNN, CNLD also has a higher PSNR, and the number of parameters is greatly reduced. Additionally, CNLD is superior compared to DnCNN in {\it Barbara}, 
which contains high-frequency components.

\section{Conclusions}
In this study, we proposed a CNLD for signal restoration. It was confirmed that image denoising by the proposed method was more efficient compared to the existing DnCNN method, especially for a small number of training samples and images containing high-frequency components.

\paragraph{Acknowledgments} This work was supported by JSPS KAKENHI Grant Number JP19H04135.


\end{document}